\relax
\documentclass[letterpaper]{article} 
\usepackage{aaai20}  
\usepackage{times}  
\usepackage{helvet} 
\usepackage{courier}  
\usepackage[hyphens]{url}  
\usepackage{graphicx} 
\usepackage{graphicx}  
\usepackage{multirow} 
\usepackage{here}
\title{SMILES Transformer: Pre-trained Molecular Fingerprint \\for Low Data Drug Discovery}
\author{Shion Honda\textsuperscript{\rm 1,2,3}, Shoi Shi\textsuperscript{\rm 1,2,3}, Hiroki R. Ueda\textsuperscript{\rm 1,2,3}\\
\textsuperscript{\rm 1}University of Tokyo\\ 
\textsuperscript{\rm 2}International Research Center for Neurointelligence\\
\textsuperscript{\rm 3}RIKEN Center for Biosystems Dynamics Research\\
shion\_honda@ipc.i.u-tokyo.ac.jp, \{sshoi0322-tky,uedah-tky\}@umin.ac.jp 
}
\begin{document}

\maketitle

\begin{abstract}
In drug-discovery-related tasks such as virtual screening, machine learning is emerging as a promising way to predict molecular properties. Conventionally, molecular fingerprints (numerical representations of molecules) are calculated through rule-based algorithms that map molecules to a sparse discrete space. However, these algorithms perform poorly for shallow prediction models or small datasets. To address this issue, we present SMILES Transformer. Inspired by Transformer and pre-trained language models from natural language processing, SMILES Transformer learns molecular fingerprints through unsupervised pre-training of the sequence-to-sequence language model using a huge corpus of SMILES, a text representation system for molecules. We performed benchmarks on 10 datasets against existing fingerprints and graph-based methods and demonstrated the superiority of the proposed algorithms in small-data settings where pre-training facilitated good generalization. Moreover, we define a novel metric to concurrently measure model accuracy and data efficiency.
\end{abstract}

\section{Introduction}
\noindent Recently, deep learning has emerged as a powerful machine learning technology. When applied to big data, deep learning can show equal or even better performance than humans in many domains such as computer vision \cite{resnet}, natural language processing (NLP) \cite{bert,xlnet}, making decisions \cite{alpha}, and medicine \cite{jtvae}. Based on projected performance benchmarks, deep learning is expected to be useful a tool to handle time-consuming tasks. 

Drug discovery is a process to find a new drug for a disease of interest from a chemical library and validate its efficacy and safety in clinical trials. This process usually takes more than a decade and is costly, and therefore may be improvable by deep learning methods. Indeed, deep learning has been applied to the process of drug discovery including quantitative structure-property relationships (QSPR) prediction \cite{graphconv,s2sfp}, molecule generation and lead optimization \cite{gomez,jtvae}, retrosynthesis planning \cite{segler,mt}, and compound-protein affinity prediction \cite{deepdta}. 

In order to apply machine learning to drug discovery, molecular data must be transformed into a readable format for machine learning. One major approach is to transform molecular data into a simplified molecular input line entry system (SMILES), a text representation of molecules that is commonly used in many databases \cite{s2sfp,gomez}. Recently, graph-based approaches \cite{graphconv,weave} have been proposed, which usually show better performance than text-based approaches, such as SMILES, in QSPR tasks. In these studies, the models are designed for large fully-labeled training data settings, which requires huge labeled datasets and a QSPR model for one-shot learning \cite{oneshot}. However, in most cases, it is difficult to prepare large labeled datasets of experimentally validated molecular properties or affinities to proteins, so that graph-based approaches might have limited application. Therefore, the development of a high-performing algorithm for small datasets will be required. 

Given recent progress in the NLP field \cite{elmo,bert,xlnet}, a pre-training approach may be a promising way to address this challenge. Language model pre-training can exploit huge unlabeled corpora to learn the representations of words and sentences and then the pre-trained model is fine-tuned to downstream tasks using a relatively smaller set of labeled data. Indeed, pre-training approaches have been implemented in the cheminformatics field: a pre-trained sequence-to-sequence learning models (seq2seq) composed of RNNs \cite{seq2seq} or variational autoencoders (VAE) \cite{vae} by decoding SMILES from the learned representations \cite{gomez,s2sfp,grammar,chemnet,bjerrum,winter}. However, these studies did not demonstrate performance in small data settings. In other words, the performance on small data settings of pre-training approaches in the cheminformatics field has not been evaluated yet. In this study, by applying the latest pre-training method in the NLP field to cheminformatics, we propose a new approach called SMILES Transformer (ST) that shows higher performance on small data settings than other approaches. ST is based on a Transformer \cite{transformer} pre-trained in an unsupervised way that produce continuous, data-driven fingerprints of molecules given SMILES. These fingerprints grasp the semantics of molecules and can be fed to arbitrary predictive models for many downstream tasks. 

\begin{figure}[t]
			\includegraphics[width=0.5\textwidth]{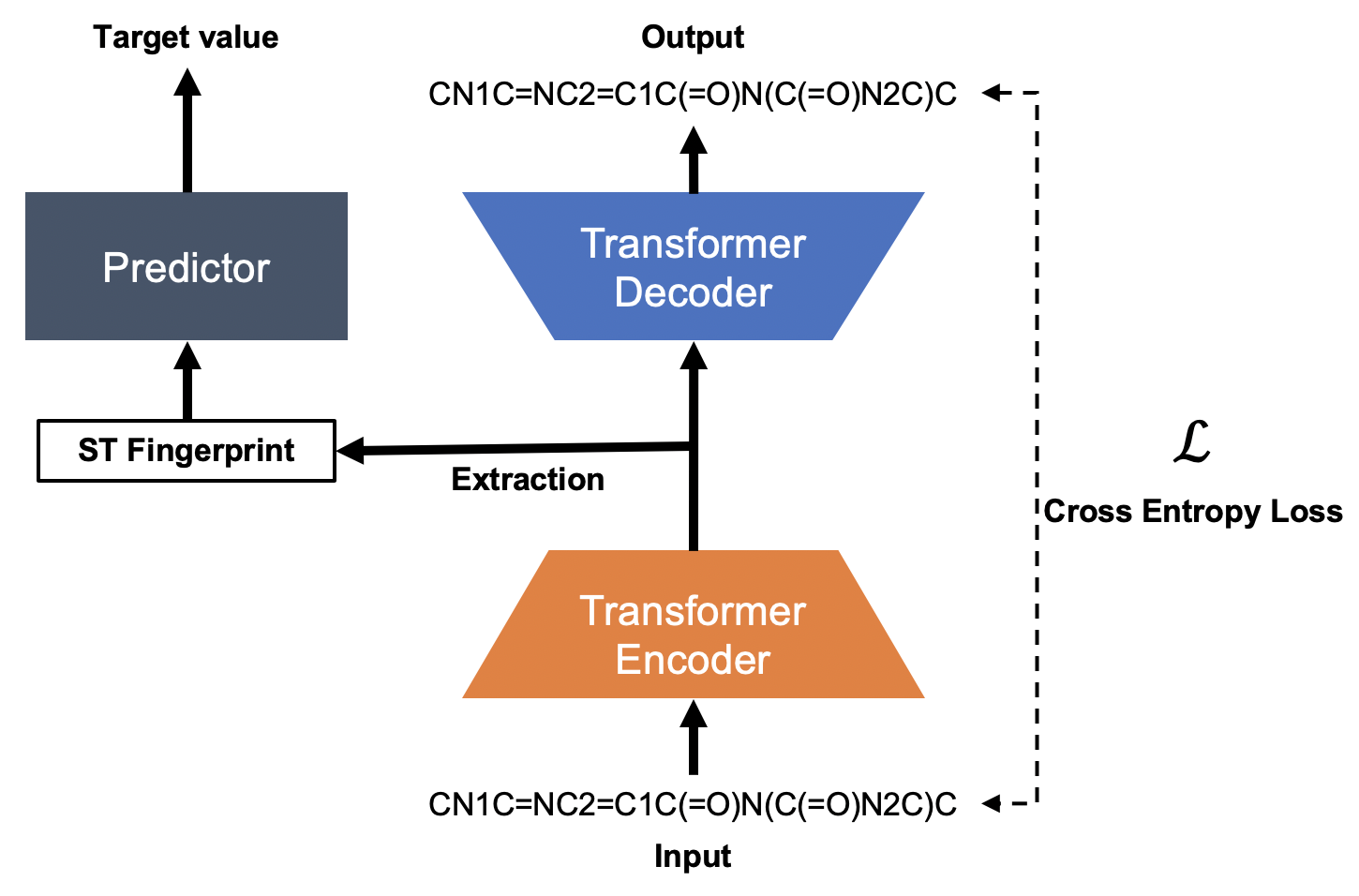}
			\caption{The illustration of SMILES Transformer pre-training and fingerprint extraction.}
			\label{abstruct}
\end{figure}

In order to evaluate the QSPR performance on small data settings, we focused on data efficiency. However, because there are few works focusing on data efficiency, which metric should be used is elusive. The most related work may be done by \cite{molnet}, where model performance is evaluated against the size of the training set and data efficiency is emphasized as well as the best score. In this study, we propose a novel scalar metric to evaluate data efficiency. Our proposed model is described in Figure \ref{abstruct}.

To sum up, our contributions include the following:
\begin{itemize}
\item We propose a data-driven fingerprinting model, SMILES Transformer, which works well with simple predictors and enables state-of-the-art data efficiency in 5 out of 10 datasets in MoleculeNet.
\item We pre-train Transformers with unlabeled SMILES to learn their representations and show the potential of text-based models compared to baseline models including graph convolutions.
\item We propose a scalar metric for data efficiency that measures model performance under different sizes of training data.
\end{itemize}

In the first section, we will explain how ST is trained and the fingerprints are extracted. In the second section, we define the metric for data efficiency. In the third section, we will compare the performance of ST fingerprints against other methods using 10 different datasets from MoleculeNet and more deeply inspect the pre-trained ST including latent space visualization. Finally, we discuss possible future directions.

\section{Methods}
\noindent In this section, we introduce the SMILES Transformer architecture, pre-training settings, and how to design ST fingerprints. We then propose a novel metric for data efficiency.

\subsection{SMILES Transformer}
\subsubsection{Model Architecture}
Unlike RNNs, Transformers \cite{transformer} do not have recurrent connections and are therefore more stable and faster to converge. Moreover, they  empirically show better featurization performance on long sequences and complicated problems than RNNs. Hence, they are chosen as the \textit{de facto} standard models in NLP \cite{bert,xlnet}.

We built an encoder-decoder network with 4 Transformer blocks for each with PyTorch \cite{pytorch}. Each Transformer block has 4-head attentions with 256 embedding dimensions and 2 linear layers. 

\subsubsection{Pre-training settings}
We pre-trained ST with 861,000 unlabeled SMILES randomly sampled from ChEMBL24, a dataset of bioactive and real molecules \cite{chembl}. The SMILES was split into symbols (e.g., 'c', 'Br', '=', '(', '2') and then the symbols were one-hot encoded to input to the network. To alleviate bias for the canonical representation of SMILES, we randomly transformed them every time they were used by the SMILES enumerator \cite{enumerate}. Following the original paper \cite{transformer}, we used the sum of token encoding and positional encoding to input to the network. The network was trained for 5 epochs to minimize the cross entropy between the input SMILES and the output probability by the Adam optimizer \cite{adam}. After convergence, the network achieved a perplexity of 1.0, meaning perfect decoding from the encoded representations.

\begin{table*}[t]
\centering
\caption{Summarized information of MoleculeNet \cite{molnet}. "R" and "C" in the type column indicates regression and classification respectively.\\}
\label{molnet}
\begin{tabular}{ccccccc}
Category                            & Dataset       & Tasks & Type      & Mols & Metric   & Description\\ 
\hline
\multirow{3}{*}{\begin{tabular}{c}
Physical \\ chemistry
\end{tabular}} & ESOL          & 1     & R     & 1128      & RMSE     & Aqueous solubility\\
                                    & FreeSolv      & 1     & R     & 643       & RMSE     & Hydration free energy\\
                                    & Lipophilicity & 1     & R     & 4200      & RMSE     & Octanol/water distribution coefficient (logD)\\ 
\hline
\multirow{3}{*}{Biophysics}         & MUV           & 17    & C & 93 127    & PRC-AUC  & 17 tasks from PubChem BioAssay\\
                                    & HIV           & 1     & C & 41 913    & ROC-AUC  & Ability to inhibit HIV replication\\
                                    & BACE          & 1     & C & 1522      & ROC-AUC  & Binding results for inhibitors of human BACE-1\\ 
\hline
\multirow{4}{*}{Physiology}         & BBBP          & 1     & C & 2053      & ROC-AUC  & Blood-brain barrier penetration\\
                                    & Tox21         & 12    & C & 8014      & ROC-AUC  & Toxicity measurements\\
                                    & SIDER         & 27    & C & 1427      & ROC-AUC  & Adverse drug reactions on 27 system organs\\
                                    & ClinTox       & 2     & C & 1491      & ROC-AUC & Clinical trial toxicity and FDA approval status 
\end{tabular}
\end{table*}

\begin{table*}[t]
\centering
\caption{Comparison of data efficiency metric (DEM) with the baseline models on 10 datasets from MoleculeNet \cite{molnet}. The up/down arrows show that the higher/lower score is better, respectively.\\}
\label{full_result}
\begin{tabular}{c|cccccccccc}
Dataset       & ESOL $\downarrow$           & FrSlv $\downarrow$            & Lipo $\downarrow$         & MUV $\uparrow$           & HIV $\uparrow$            & BACE $\uparrow$           & BBBP $\uparrow$          & Tox21 $\uparrow$         & Sider $\uparrow$         & ClinTox $\uparrow$        \\
\hline
ST+MLP (Ours) & \textbf{1.144} & \textbf{2.246} & 1.169          & 0.009          & 0.683          & 0.719          & \textbf{0.900} & \textbf{0.706} & 0.559          & \textbf{0.963}  \\
ECFP+MLP      & 1.741          & 3.043          & 1.090          & \textbf{0.036} & 0.697          & \textbf{0.769} & 0.760          & 0.616          & \textbf{0.588} & 0.515           \\
RNNS2S+MLP    & 1.317          & 2.987          & 1.219          & 0.010          & 0.682          & 0.717          & 0.884          & 0.702          & 0.558          & 0.904           \\
GraphConv     & 1.673          & 3.476          & \textbf{1.062} & 0.004          & \textbf{0.723} & 0.744          & 0.795          & 0.687          & 0.557          & 0.936          
\end{tabular}
\end{table*}

\subsubsection{Fingerprint extraction}
As the outputs of the Transformers are contextualized word-level representations, ST outputs a sequence of symbol-level (atom-level) representations. Therefore, we need to pool them to obtain the molecule-level representations (fingerprints). We concatenated the four vectors to get the fingerprints: mean and max pooled output of the last layer, the first output of the last and the penultimate layer. Now we have a 1024-dimensional fingerprint for each molecule from ST. This fingerprint is designed to have the same dimensionality with the baseline we use for, the extended-connectivity fingerprint (ECFP) \cite{ecfp}.

\subsection{Data Efficiency Metric (DEM)}
Here we discuss how to measure the data efficiency of a predictive model $f$ in terms of the metric $m$. Intuitively, data efficiency can be measured by averaging the metric $m$ of the model $f$ trained with different sizes of the training data.

More formally, let $(X,Y)$ denote the whole available dataset and $(X_i,Y_i)$ denote the test data sampled from $(X,Y)$ at the rate of $1-i$. Then, the training data and the model trained with them can be represented as $(X\setminus X_i, Y\setminus Y_i)$ and $f_i$, respectively. The metric $m$ should be chosen to be suitable for the tasks. That is, in classification tasks $m$ should be the area under the receiver operation characteristics (ROC-AUC) or the F1 score and in regression tasks $m$ should be the R2 score or the root mean squared error (RMSE). 
 
Now the proposed Data Efficiency Metric (DEM) is formulated as: 

\begin{equation}
M_{DE}(f,m) = \frac{1}{|I|}\sum_{i\in I}m(f_i, X_i, Y_i)
\end{equation}

Since we used various datasets with a wide range of sizes in the experiment described below, the percentage of the training data $i$ should be increased exponentially. Therefore, $i$ is doubly increased from 1.25\% to 80\%, i.e., $I=\{0.0125,0.025,...,0.4,0.8\}$.

\section{Experiments}
\noindent We conducted five experiments to see how SMILES Transformer works from different perspectives. First, we evaluated the performance of ST against other baseline models on 10 chemical datasets. Second, we visualized the latent space to answer the question: why do ST fingerprints work well for certain datasets? Third, we applied linear models to ST and other fingerprints in order to validate that ST maps molecules to a good latent space by minimizing the contribution of the models themselves. Fourth, we evaluated our ST and baseline models on a stratified dataset by the lengths of SMILES to see when ST provides an advantage. Finally, we compared the maximum performance of ST against state-of-the-art models under large data settings.

\subsection{Performance on Downstream Tasks}
\subsubsection{Datasets}
We evaluated the performance of our pre-trained SMILES Transformer on 10 datasets from MoleculeNet \cite{molnet}, a benchmark for molecular property prediction. These datasets were chosen because they do not use 3D information and the sizes are not too large. The datasets are different from each other in their domains, task types, and sizes.

\begin{itemize}
    \item Physical chemistry: ESOL, FreeSolv, and Lipophilicity
    \item Biophysics: MUV, HIV, and BACE
    \item Physiology: BBBP, Tox21, SIDER, and ClinTox
\end{itemize}

The information about each dataset is summarized in Table \ref{molnet}. For the evaluation metrics, we used the root mean squared error (RMSE) for the regression tasks and the area under the receiver operating characteristic curve (ROC-AUC) or the area under the precision-recall curve (PRC-AUC) for the classification tasks as suggested in \cite{molnet}.

\subsubsection{Baseline models}
We compared our pre-trained SMILES Transformer to the following three baseline models for molecular property prediction tasks:
\begin{itemize}
\item ECFP4 \cite{ecfp} is a hand-crafted fingerprint. It hashes multi-scaled substructures to integers and makes a fixed-length binary vector where 1 indicates the existence of the assigned substructure and 0 for the absence. ECFP4 counts substructures with the diameters up to 4.
\item RNNS2S \cite{s2sfp} is another text-based pre-trained fingerprint that adopts RNN Seq2seq for the model architecture.
\item GraphConv \cite{graphconv} learns and predicts the target value directly through graph convolution operations, rather than extracting fingerprints and building another model for supervised downstream tasks. Although GraphConv is not a task-agnostic fingerprint, we include it here as the state-of-the-art model.
\end{itemize}

We used RDKit \cite{rdkit} to compute ECFP4 and DeepChem \cite{deepchem} implementation of GraphConv (with the default hyperparmeters). For RNNS2S, we implemented it with PyTorch \cite{pytorch} and pre-trained it with the same dataset as ST. The encoder and the decoder are both 3-layer bidirectional gated recurrent units (GRUs) \cite{gru} with 256 hidden vector dimensions. We obtained the same dimension of fingerprint as ST by concatenating two outputs from the last and the penultimate layer.

\subsubsection{Experiment settings}
In the downstream tasks, we used simple models, such as multilayer perceptron (MLP) classifiers and regressors with the same default hyperparameters in scikit-learn \cite{sklearn} in order to evaluate the performance of the three fingerprints, themselves as much as possible. All of these fingerprints have 1,024 dimensions. The datasets were randomly split (stratified for classification) to train sets and test sets by the percentage $i$. Note that we did not use a scaffold split suggested in \cite{molnet}. We ran 20 trials for each split and report the mean score and standard deviation in Figure \ref{results} and DEM in Table \ref{full_result}. The metrics were chosen as recommended in MoleculeNet.

\subsubsection{Results}
Table \ref{full_result} shows DEM of the 4 models. ST achieves the best score in 5 out of 10 datasets, followed by ECFP and GraphConv. 

\begin{figure*}[p]
			\centering
			\includegraphics[width=1.0\textwidth]{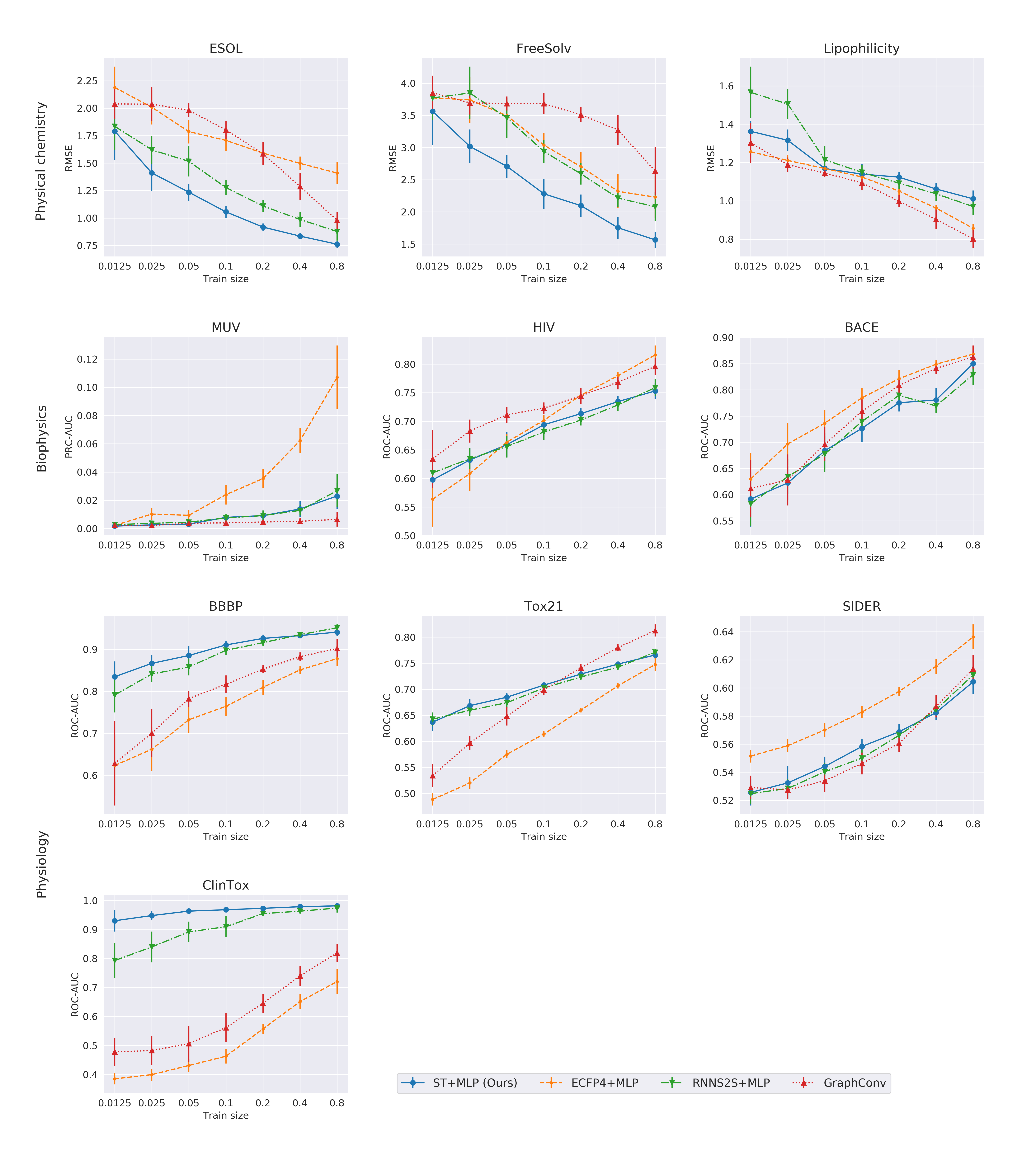}
			\caption{Comparison of model performance against different train size on the 10 datasets. The top row indicates the results for the physical chemistry datasets, the second row indicates biophysics, and the two bottom rows indicate physiology, respectively. The scores were averaged over 20 trials and the error bars are the standard deviations}
			\label{results}
\end{figure*}

See Figure \ref{results} for the performance change against the train size. In ESOL, FreeSolv, BBBP, and ClinTox, ST performs the best at almost all points by a significant margin and especially high scores when the train size is small compared to the other models. In Tox21, ST supports good prediction along RNNS2S, but is beaten by GraphConv as the train size increase. In Lipophilicity, MUV, BACE, and SIDER, ECFP or GraphConv can predict better than ST.

\subsection{Visualization of the Latent Space}
To inspect why our ST fingerprints lead to good predictive performance, we visualized the latent space and decode some samples from it. For each dataset in MoleculeNet, we conducted the following procedure:

\begin{enumerate}
    \item Calculate the ST fingerprint (1024-dimension) of each molecule.
    \item Reduce their dimensions to 2 with t-SNE \cite{t-sne}.
    \item Plot the reduced features into a 2-dimensional space coloring by the target value.
    \item Choose a trajectory in the 2-dimensional space and divide it into 12 points.
    \item Find the nearest neighbors of the 12 points and draw the corresponding molecules.
\end{enumerate}

\begin{figure*}[p]
			\includegraphics[width=0.85\textwidth]{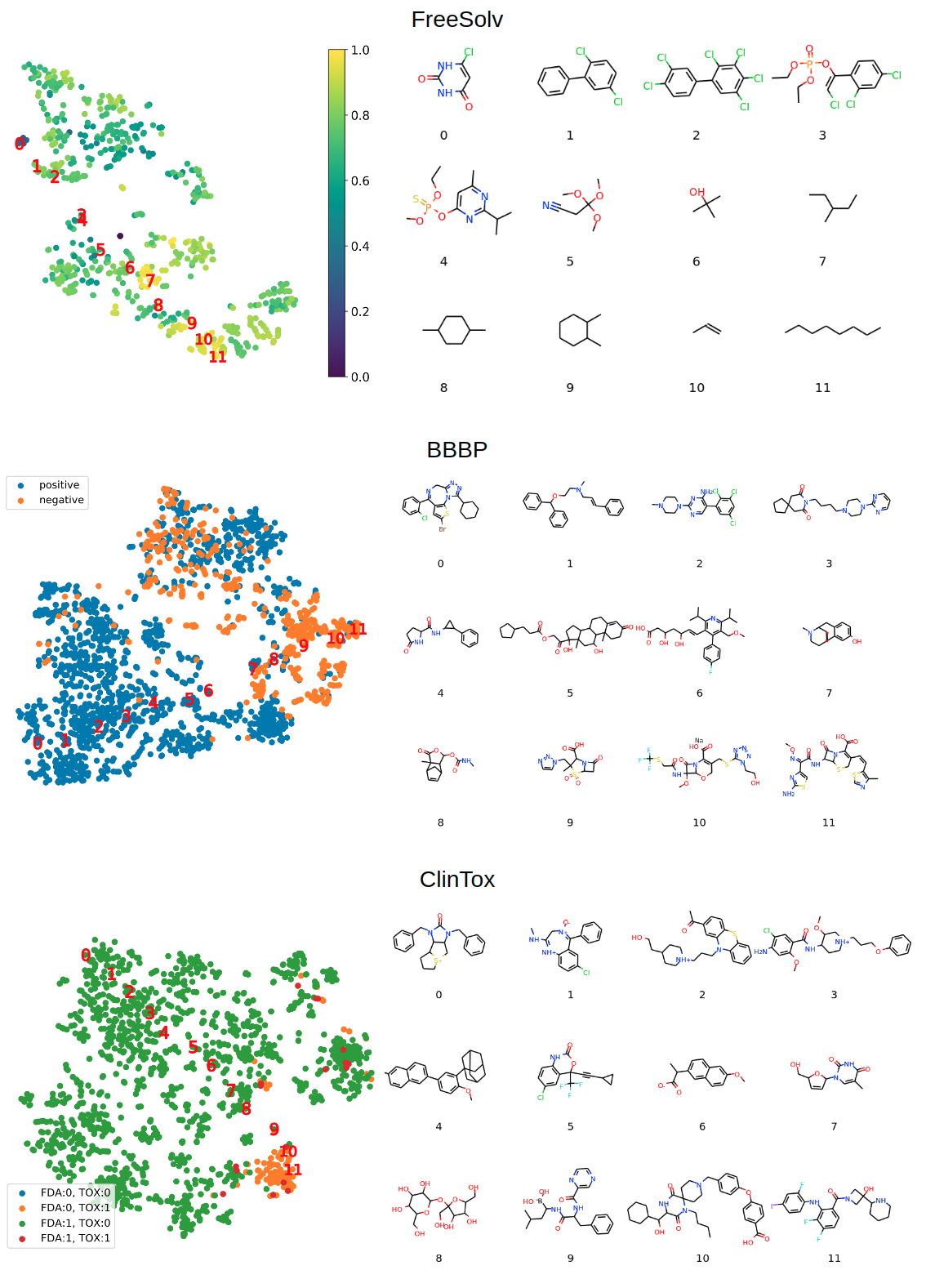}
			\caption{Visualization of the latent space of SMILES Transformer. For three datasets, FreeSolv, BBBP, and ClinTox, the dimensions of ST fingerprints of the molecules are reduced to 2 with t-SNE \cite{t-sne}. Then, the nearest neighbors of the 12 data points on a trajectories are plotted on the latent space (left panel). The 12 points are decoded to molecules and shown in the right panel. The color bar of the top left panel indicates the standardized free energy.}
			\label{tsne}
\end{figure*}

We show the result of the three datasets where ST fingerprints work especially well in Figure \ref{tsne}, that is, FreeSolv, BBBP, and ClinTox. In FreeSolv, it can be seen that there is a clear gradation from upper left to lower right, and the molecule becomes simpler (i.e., less loops and branches) along the trajectory. In BBBP and ClinTox, the categorical target values are successfully separated, but there is no clear trends in the decoded molecules.

\subsection{Application of Simple Predictive Models}
In Section 3.1, we used MLP for the predictive model to ST, RNNS2S, and ECFP, expecting that combining it with these fingerprints would work comparably or better than GraphConv. Here we used the simplest models to measure the pure effect of the fingerprints. To be specific, adding L2 regularization to avoid overfitting, we used ridge regression for regression tasks and logistic regression with L2 penalty for classification tasks. We excluded MUV and SIDER from this experiment because their highly imbalanced columns caused errors in the solver and ROC-AUC functions implemented in scikit-learn \cite{sklearn}. We followed the same procedure as in Section 3.1 except for the model selection and datasets and the results are shown in Table \ref{simple}.

\begin{table*}[ht]
\centering
\caption{Comparison of data efficiency metric (DEM) with the baseline models on the 8 datasets from MoleculeNet \cite{molnet}. The predictive models are ridge regression and logistic regression with L2 penalty. The up/down arrows show that the higher/lower score is better, respectively.\\}
\label{simple}
\begin{tabular}{c|cccccccc}
Dataset       & ESOL $\downarrow$           & FrSlv $\downarrow$            & Lipo $\downarrow$         & HIV $\uparrow$            & BACE $\uparrow$           & BBBP $\uparrow$          & Tox21 $\uparrow$         &  ClinTox $\uparrow$        \\
\hline
ST (Ours) & \textbf{1.140} & \textbf{2.452} & 1.213 & 0.696 & 0.720 & \textbf{0.895} & \textbf{0.711} & \textbf{0.958}\\
ECFP &      1.678 & 2.843 & \textbf{1.174} & \textbf{0.727} & \textbf{0.790} & 0.825 & 0.710 & 0.704\\
RNNS2S &   1.288 & 2.881 & 1.194 & 0.688 & 0.727 & 0.884 & 0.709 & 0.915      
\end{tabular}
\end{table*}

\begin{table*}[ht]
\centering
\caption{Comparison of the best achieved scores with the record scores on the 8 datasets from MoleculeNet \cite{molnet}. The scores of ECFP, GraphConv, and Weave are the reported scores in MoleculeNet. The up/down arrows show that the higher/lower score is the better, respectively.\\}
\label{record}
\begin{tabular}{c|cccccccc}
Dataset       & ESOL $\downarrow$           & FrSlv $\downarrow$            & Lipo $\downarrow$         & HIV $\uparrow$            & BACE $\uparrow$           & BBBP $\uparrow$          & Tox21 $\uparrow$         &  ClinTox $\uparrow$ \\
Splitting & random & random & random & scaffold & scaffold & scaffold & random & random\\ 
\hline
ST (Ours) & 0.72 & 1.65 & 0.921 & 0.729 & 0.701 & 0.704 & 0.802 & \textbf{0.954} \\
ECFP      & 0.99 & 1.74 & 0.799 & \textbf{0.792} & \textbf{0.867} & \textbf{0.729} & 0.822 & 0.799\\
GraphConv & 0.97 & 1.40 & \textbf{0.655} & 0.763 & 0.783 & 0.690 & \textbf{0.829} & 0.807 \\ 
Weave     & \textbf{0.61} & \textbf{1.22} & 0.715 & 0.703 & 0.806 & 0.671 & 0.820 & 0.832 \\ 
\end{tabular}
\end{table*}

Our ST fingerprints with linear models achieved the best scores in 5 out of 8 datasets, indicating that the ST fingerprint is a strong fingerprint that leads to the best performance regardless of model selection. 

\subsection{Stratified Scores by the Size of Molecules}
We conducted another study to inspect when ST has an advantage against other models. We stratified the BBBP dataset by the lengths of SMILES (similar to the sizes of the molecules) into 5 groups and evaluated within each group. The scores and the distributions of the lengths of SMILES are shown in Figure \ref{strat}.

Figure \ref{strat} indicates that the ROC-AUC score of ST increases along the length of SMILES, which is a similar trend to the other text-based fingerprint, RNNS2S. On the other hand, GraphConv shows more or less the same performance regardless of the SMILES lengths. These results suggest that longer SMILES give ST richer information for better discrimination.

\subsection{Comparison with Record Scores}
Finally, we compared the maximum performance of ST under the large data setting with the reported scores in MoleculeNet. Since the ST fingerprint is proven to be better than the RNNS2S fingerprint, we omitted it and instead added another graph-based model named Weave \cite{weave} to the baselines. In this experiment, the datasets were split into train, validation, test sets with the proportion of 80\%, 10\%, 10\%. The validation sets were used for hyperparameter tuning and the test sets were only used for calculating the scores. To fairly compare with the reported scores, the datasets HIV, BACE, BBBP used a scaffold split and the others were split randomly. We choose the model and hyperparameter set achieving the best validation score with optuna \cite{optuna}, from a linear model with L2 penalty, MLP, and LightGBM \cite{lgbm}. We conducted three independent runs and reported the average scores in Table \ref{record}

ST achieves first place only in ClinTox, but performs comparable to ECFP and graph-based models in the other datasets. We can conclude that our ST fingerprints, if carefully tuned, are still useful even when the large number of labels are available. 

\section{Conclusions}
In this paper, we propose SMILES-Transformer, a data-driven molecular fingerprint produced by a Transformer-based seq2seq pre-trained with a huge set of unlabeled SMILES. ST fingerprints were shown to work well with any predictive model in MoleculeNet downstream tasks and is effective especially when there is not enough labeled data. When large labeled data are available, ST fingerprints work comparable to other state-of-the-art baselines such as GraphConv. We also propose DEM, a novel metric for data efficiency. In terms of DEM, the ST fingerprint is better than existing methods in 5 out of 10 downstream tasks.

Future work can continue in three directions. First, replacing the Transformer in ST with Transformer-XL, an extended model that can handle much longer sequences, will alleviate the length limit of ST. Second, ST will be even stronger when trained in a multi-task fashion as done in ChemNet \cite{chemnet}: predicting automatically-calculated molecular descriptors (e.g., molecular weight, LogP) as well as decoding the input SMILES. This will help the model to learn more chemistry-relevant representations. Finally, making use of the information of enumerated SMILES is one of the keys to improving text-based molecular representations. As done in \cite{bjerrum}, a set of different SMILES of the same molecule can be used to restrict the latent space.

Our implementation for SMILES-Transformer is available at \url{https://github.com/DSPsleeporg/smiles-transformer}

{
\vspace{10pt}
\begin{figure}[hp]
            \centering
			\includegraphics[width=0.3\textwidth]{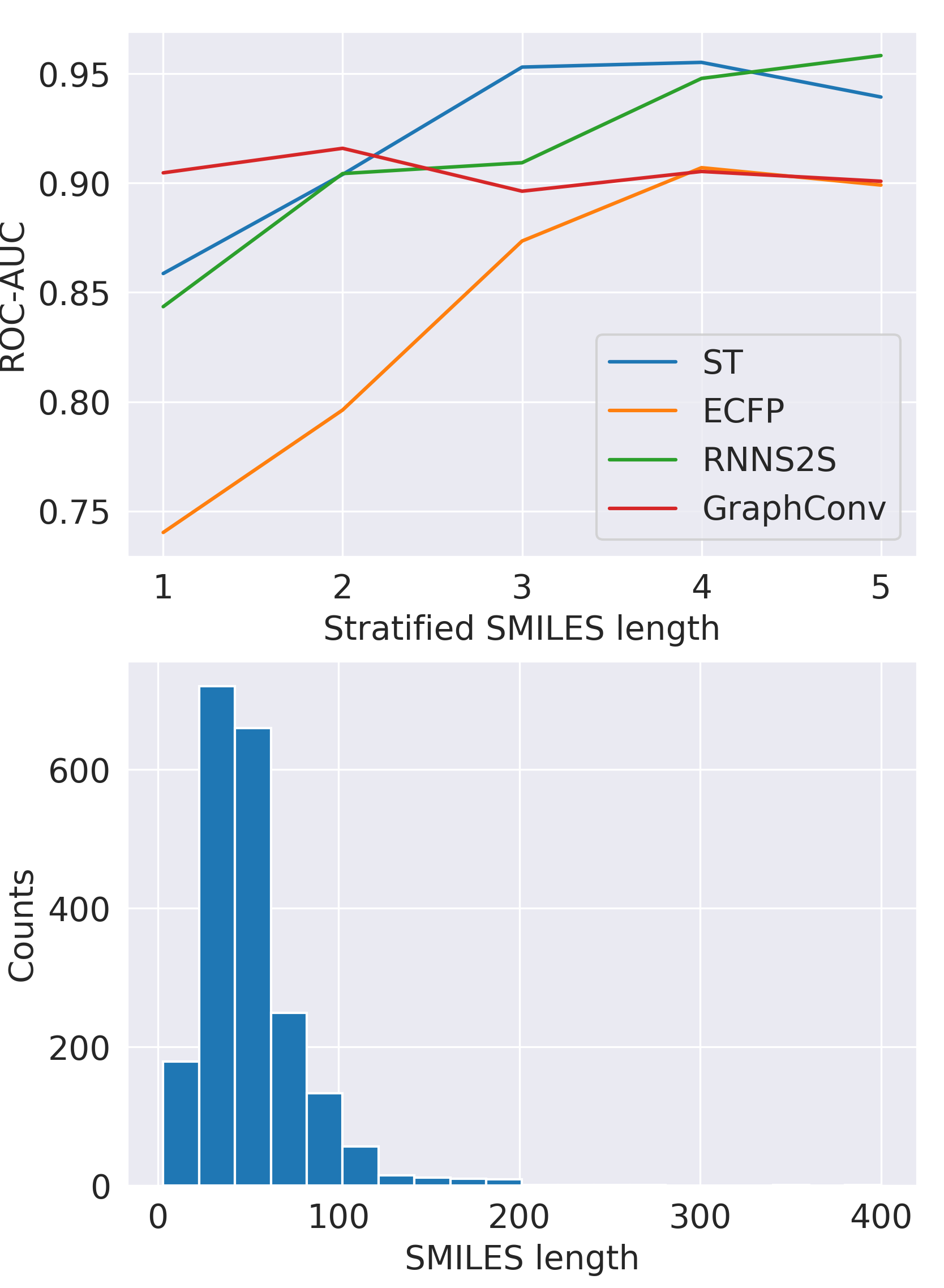}
			\caption{ROC-AUC scores on each stratified group by the lengths of SMILES (left) and the distributions of the lengths of SMILES (right) of BBBP dataset.}
			\label{strat}
\end{figure}
}

\clearpage
\bibliography{bib.bib}
\bibliographystyle{aaai}

\end{document}